\documentclass[conference,a4paper]{IEEEtran}
\IEEEoverridecommandlockouts
\usepackage{cite}
\usepackage{amsmath,amssymb,amsfonts}
\usepackage{algorithmic}
\usepackage{graphicx}
\ifCLASSOPTIONcompsoc
    \usepackage[caption=false, font=normalsize, labelfont=sf, textfont=sf]{subfig}
\else
\usepackage[caption=false, font=footnotesize]{subfig}
\fi
\usepackage{textcomp}
\usepackage{hyperref}
\usepackage{url}
\hypersetup{urlcolor=blue, colorlinks=true}
\usepackage{xcolor}

\def\BibTeX{{\rm B\kern-.05em{\sc i\kern-.025em b}\kern-.08em
    T\kern-.1667em\lower.7ex\hbox{E}\kern-.125emX}}
\begin{document}

\title{Underwater Color Restoration Using U-Net Denoising Autoencoder}

\author{\IEEEauthorblockN{Yousif Hashisho}
\IEEEauthorblockA{\textit{Department of Maritime Graphics} \\
\textit{Fraunhofer Institute for Computer Graphics Research (IGD)}\\
Rostock, Germany \\
yousif.hashisho@igd-r.fraunhofer.de}\\
\IEEEauthorblockN{Tom Krause}
\IEEEauthorblockA{\textit{Department of Maritime Graphics} \\
\textit{Fraunhofer Institute for Computer Graphics Research (IGD)}\\
Rostock, Germany \\
tom.krause@igd-r.fraunhofer.de}
\and
\IEEEauthorblockN{Mohamad Albadawi}
\IEEEauthorblockA{\textit{Department of Maritime Graphics} \\
\textit{Fraunhofer Institute for Computer Graphics Research (IGD)}\\
Rostock, Germany \\
mohamad.albadawi@igd-r.fraunhofer.de}\\
\IEEEauthorblockN{Uwe Freiherr von Lukas}
\IEEEauthorblockA{\textit{Department of Maritime Graphics} \\
\textit{Fraunhofer Institute for Computer Graphics Research (IGD)}\\
\textit{Department of Computer Science} \\
\textit{University of Rostock}\\
Rostock, Germany \\
uwe.freiherr.von.lukas@igd-r.fraunhofer.de}
}

\maketitle

\begin{abstract}
Visual inspection of underwater structures by vehicles, e.g. remotely operated vehicles (ROVs), plays an important role in scientific, military, and commercial sectors. However, the automatic extraction of information using software tools is hindered by the characteristics of water which degrade the quality of captured videos. As a contribution for restoring the color of underwater images, Underwater Denoising Autoencoder (UDAE) model is developed using a denoising autoencoder with U-Net architecture. The proposed network takes into consideration the accuracy and the computation cost to enable real-time implementation on underwater visual tasks using end-to-end autoencoder network. Underwater vehicles perception is improved by reconstructing captured frames; hence obtaining better performance in underwater tasks. Related learning methods use generative adversarial networks (GANs) to generate color corrected underwater images, and to our knowledge this paper is the first to deal with a single autoencoder capable of producing same or better results. Moreover, image pairs are constructed for training the proposed network, where it is hard to obtain such dataset from underwater scenery. At the end, the proposed model is compared to a state-of-the-art method. 
\end{abstract}

\begin{IEEEkeywords}
autoencoders, underwater image, image restoration, Generative Adverarial Networks, real-time
\end{IEEEkeywords}

\section{Introduction}
Marine robots, such as remotely operated vehicles (ROVs), are being increasingly used in the scientific, military, and commercial sectors. They are critical in collecting data and performing certain underwater operations. Due to safety and health concerns, human intervention can be risky and limited when executing underwater missions~\cite{wynn2014autonomous}. Thus, underwater vehicles are supplied with cameras systems for performing numerous vision tasks. For instance, Choi et al., 2017~\cite{choi2017development} operated an ROV manually for inspecting harbour structures and acquiring high quality videos. Manjunatha et al., 2018~\cite{manjunatha2018low} built a robot equipped with a high definition camera for visual inspection at a specified depth in a water body. However, the automatic extraction of information using software tools is hindered by underwater image degradation caused by poor water medium and light behaviour. 

Contrast loss and color distortion affect the algorithms and ultimately the vehicle performance in gathering data and processing them. An image enhancement technique is needed for vehicle navigation by human operator to facilitate underwater tasks. Furthermore, the processing speed should be taken into consideration for a real-time implementation. 

This paper proposes Underwater Denoising Autoencoder (UDAE), a deep learning network based on a single denoising autoencoder~\cite{vincent2008extracting} using U-Net~\cite{ronneberger2015u} as a CNN architecture, for improving the quality of underwater imagery and video material. The contributions presented in this paper can be summarized as follows:
\begin{itemize}
\item UDAE network is proposed which is specialized in underwater color restoration.
\item Faster processing speed is achieved than the state-of-the-art method which optimize the real-time capability. 
\item A new dataset with a combination of different underwater scenarios (turbidity, depth, temperature, attenuation type..) is synthesized for training the proposed network. The synthetic dataset is generated using a generative deep learning method. 
\item The fully end-to-end proposed model generalizes well (real underwater images) with different degradation types. 
\end{itemize}

The rest of the paper is as follows: \S \ref{RelatedWork} talks about relevant work; \S \ref{Methodology} gives experiments and methods followed to restore underwater images; \S\ref{Results} presents corrected underwater images and the performance of the proposed network; finally, \S\ref{Conclusion} summarizes the paper.

\section{Related Work}\label{RelatedWork}

Numerous attempts have been made with different image improvement methods for restoring the color of raw underwater images. These methods fall into two categories~\cite{lu2017underwater}: hardware-based methods~\cite{schechner2005recovery,treibitz2009active} and software-based methods~\cite{ancuti2018color,farhadifard2015learning,chiang2012underwater}. Software-based methods invert the formation of underwater images and construct physical models for image enhancement in addition to modifying the image pixel values. Hardware-based methods capture multiple images with help of polarization filters, stereo setups or specialized hardware devices and use the obtained additional information~\cite{schettini2010underwater,lu2017underwater}.

Both categories show good performance, however, they are limited to certain scenarios and don't match various underwater lightening conditions. They are expensive to implement since some of them use specialized sensors and multiple images for the enhancement. Recent approaches have focused on Generative Adversarial Networks (GANs) as a new way for achieving better results.

When improving underwater imagery using deep learning models (e.g. GANs), image pairs consisting of clean and distorted underwater images are needed for training the model. It is hard to capture clean underwater images without the attenuation of light and other underwater effects. Thus, several works have been done to synthesize training images.

\begin{figure}[htbp]
	\centerline{\includegraphics[width=1.0\columnwidth]{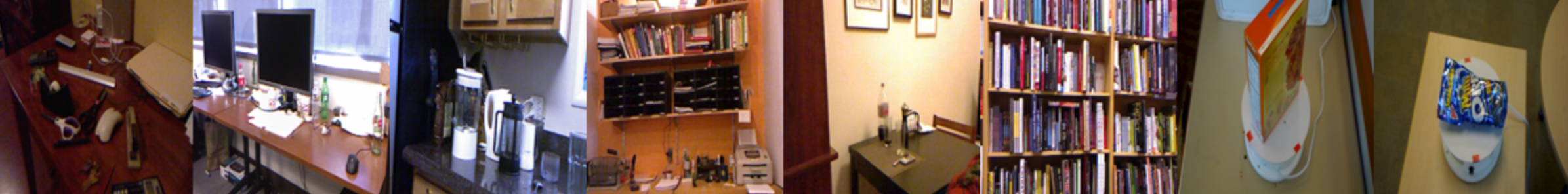}}
	\centerline{\includegraphics[width=1.0\columnwidth]{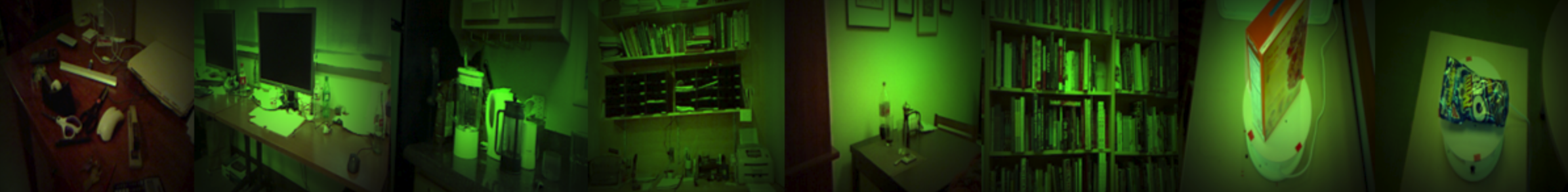}}
	\caption{The synthesized underwater images by WaterGAN. The color restoration might be limited to certain types of color degradation appearance.~\cite{li2018watergan}}
	\label{fig:waterGAN}
\end{figure}

Li et al., 2018~\cite{li2018watergan} used two types of networks: Water Generative Adversarial Network (WaterGAN) for generating realistic underwater images and Underwater Image Restoration Network for correcting the color. The generator of WaterGAN models the formation of underwater image using three stages: Attenuation, Scattering, and Camera Model. After that, the learned generator is used to generate training image samples for the color restoration network. First, a relative depth map is estimated and reconstructed from the input image and are both used for color restoration. They showed efficiency for real-time applications, however, their network is limited to certain degradation type appearance due to the way of generating images. Figure~\ref{fig:waterGAN} shows the images that were used for training the network which do not reflect underwater structures. The clean images consist of in-air images, whereas the corruption process is limited to certain degradation types (e.g. greenish mask).
\\

As an improvement over the aforementioned data generation method, Li et al., 2018~\cite{li2018emerging} and Fabbri et al., 2018~\cite{fabbri2018enhancing} used CycleGAN~\cite{zhu2017unpaired} for generating underwater images. After synthesizing the data, it was later used for training their color restoration model. 

The previously mentioned deep learning methods showed good performance in restoring the color. However in certain scenarios, they led to an unrealistic color correction of underwater images as in Li et al., 2018~\cite{li2018emerging}. The training dataset lacked true colors of underwater structures such as coral reefs and fish. Furthermore, a drawback in the color restoration model, Underwater Generative Adversarial Network (UGAN), of Fabbri et al., 2018~\cite{fabbri2018enhancing} is the efficiency of real-time implementation with high resolution images, as the model's architecture makes it computationally costly.

We follow the same procedure as in Fabbri et al., 2018~\cite{fabbri2018enhancing} for generating synthetic images. However, a different set of images is used for the training of CycleGAN. Fabbri et al., 2018~\cite{fabbri2018enhancing} collected clear underwater images and style-transferred the characteristics of degradation from distorted underwater images to them. Our generated dataset is composed of various underwater locations with different degradation types, leading to a better generalization than their network.

\section{Methodology}\label{Methodology}

Two important aspects are discussed in building the UDAE model. The first aspect is the methodology followed to generate the underwater dataset for training the network. The second one is the architecture of the UDAE model and the benefits of using a denoising autoencoder.

\subsection{Dataset}\label{Dataset}

A dataset is gathered and filtered to be used for the generation process of the image pairs. This section is divided into two subsections. The first subsection discusses data collection of underwater images, while the second discusses generating data for obtaining underwater image pairs. 

\subsubsection{Data Collection}

To train a network capable of restoring the true underwater color from the distorted images, clear images were gathered without light scattering in them. These images were taken from different sources on the Internet. As it is hard to get clear images, it was possible to obtain them from:

\begin{itemize}
\item Large fish aquariums such as the ones in museums and touristic towers.
\item Underwater images that were captured in a close distance to the structures with artificial light exposure.
\item Various images and frames taken from videos that were enhanced and processed by commercial software tools.
\end{itemize} 

The clean images were chosen based on contrast loss and degradation presented in underwater images. After that, distorted images were gathered with different attenuation types from various locations. Some of them were captured by Fraunhofer IGD from the Baltic Sea, while the others were gathered from the Internet corresponding to different locations, depths, temperatures and other degradation factors.

\subsubsection{Image Pairs Data Generation}

\begin{figure}[htbp]
	\centering
	\subfloat[\textbf{A} (clean images samples)]{
	\includegraphics[width=25mm,height=25mm]{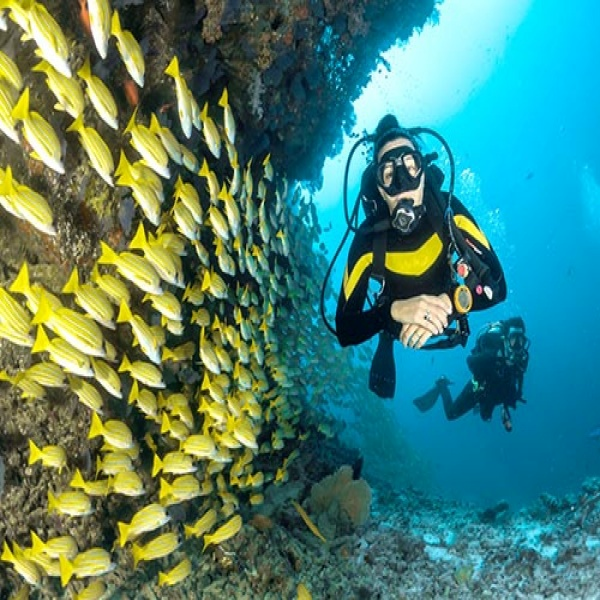}
    \includegraphics[width=25mm,height=25mm]{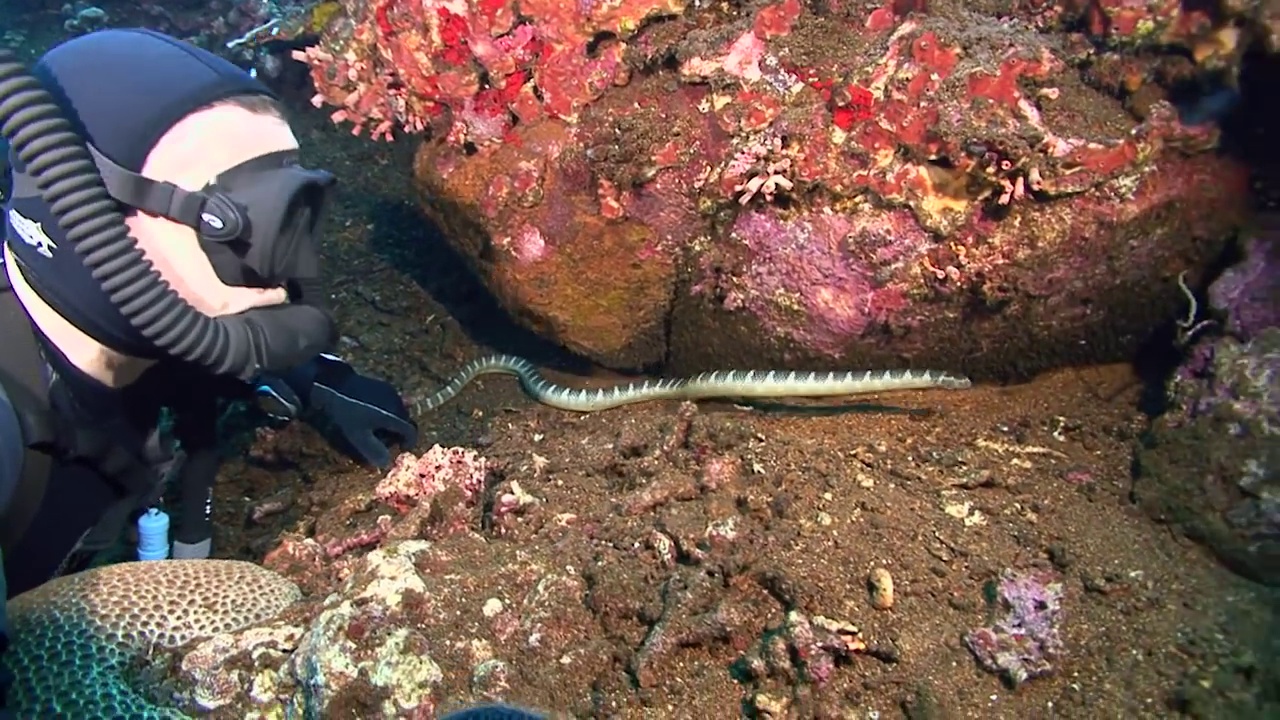}}
    \\
	\subfloat[\textbf{B} (distorted images samples)]{
	\includegraphics[width=25mm,height=25mm]{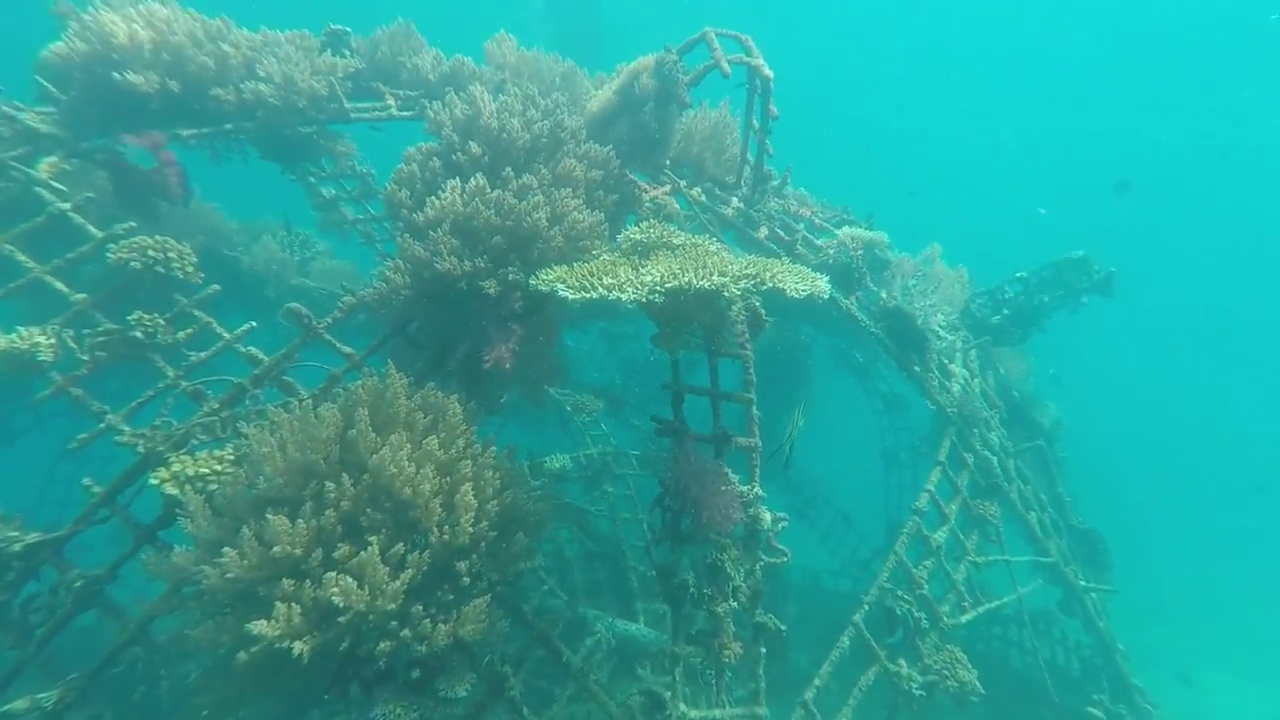}
    \includegraphics[width=25mm,height=25mm]{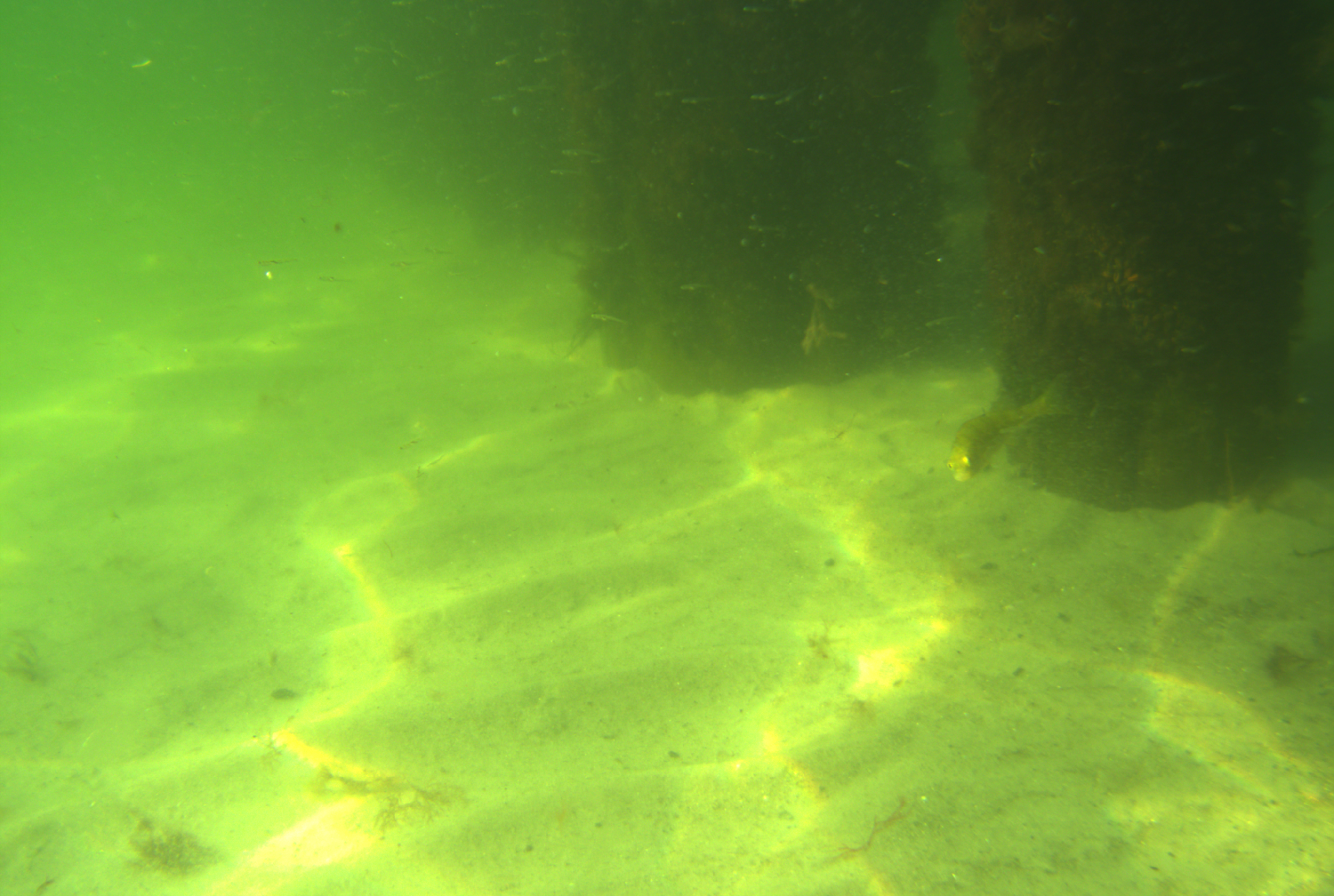}}
	\caption{Samples of images used for the style-transferring.}
	\label{fig:samples_dataset}
\end{figure}

$15,131$ images composed of clear and distorted images were collected. After that, the collected images were filtered, based on a subjective quality evaluation, into two categories: \textbf{A} (clear) containing $7,055$ images and \textbf{B} (distorted) containing $8,076$ images. The two different categories are shown in Figure~\ref{fig:samples_dataset}. All images were resized to $512 \times 512$ using area interpolation method.

\begin{figure}[htbp]
	\centering
	\subfloat[\textbf{A} (clean images samples)]{
	\includegraphics[width=25mm,height=25mm]{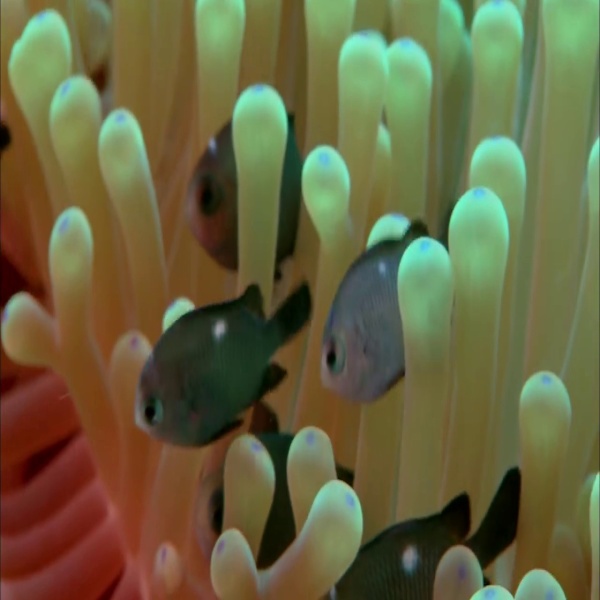}
    \includegraphics[width=25mm,height=25mm]{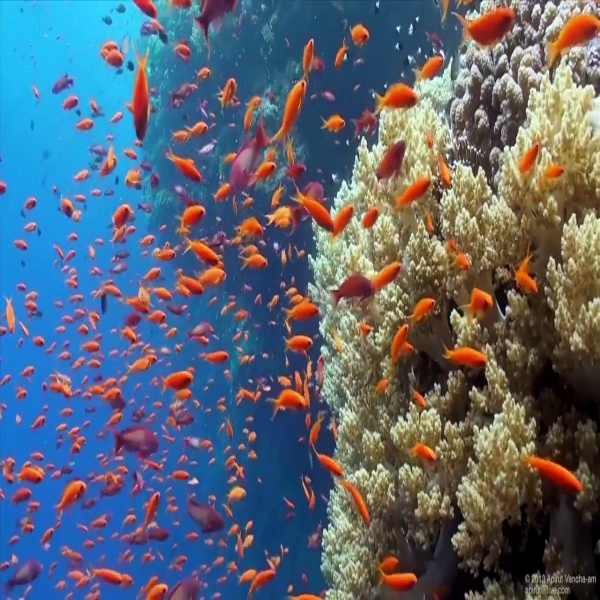}}
    \\
	\subfloat[\textbf{A}$^\prime$ (distorted images samples)]{
	\includegraphics[width=25mm,height=25mm]{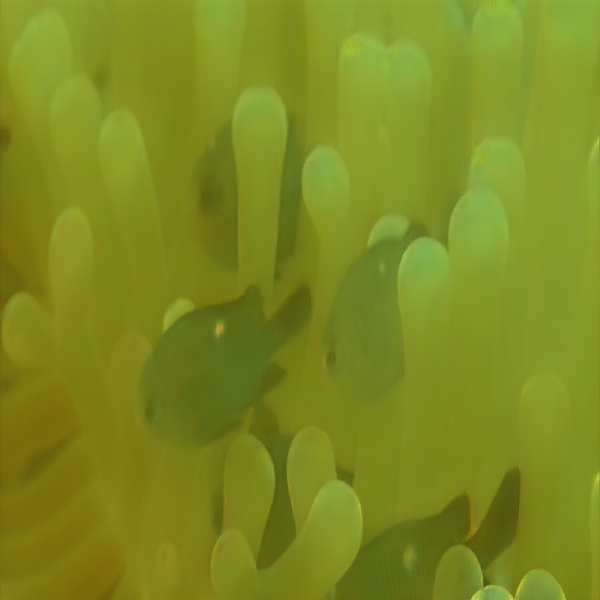}
    \includegraphics[width=25mm,height=25mm]{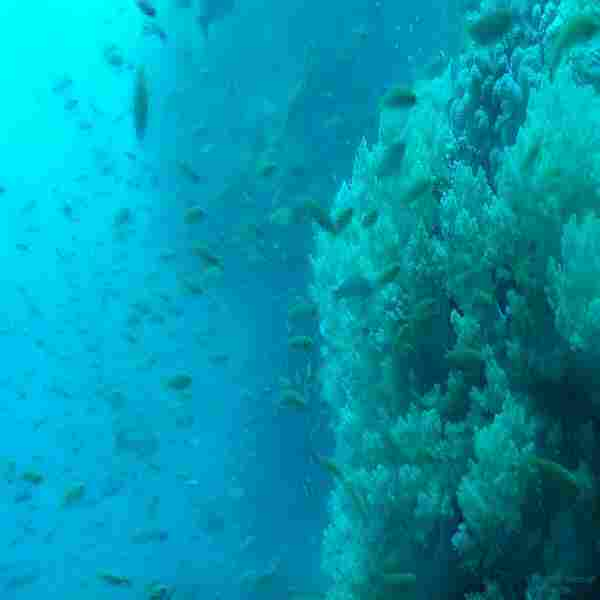}}
	\caption{Samples of the UDAE dataset after the generation of image pairs.}
	\label{fig:UDAE_samples}
\end{figure}

After gathering suitable images, CycleGAN generative model was used for style-transferring. It uses adversarial loss for learning a mapping from a source domain \textbf{X} to a target domain \textbf{Y} ($G:X\rightarrow Y$)~\cite{zhu2017unpaired}. It was used to transfer the underwater style from \textbf{B} images to \textbf{A} ones, and the result was the category \textbf{A}$^\prime$, Figure~\ref{fig:UDAE_samples}. The image pairs in \textbf{A} and \textbf{A}$^\prime$ were then used to train the autoencoder. The training of the CycleGAN model took around $9$ days on $4$ NVIDIA TITAN X GPU devices, after that $7,055$ image pairs were generated and filtered into $5,194$ after removing failure cases. The failure cases are due to limitations in the style transfer of CycleGAN.

\subsection{Proposed Network}
\begin{figure*}[htbp]
	\centerline{\includegraphics[width=1.0\textwidth]{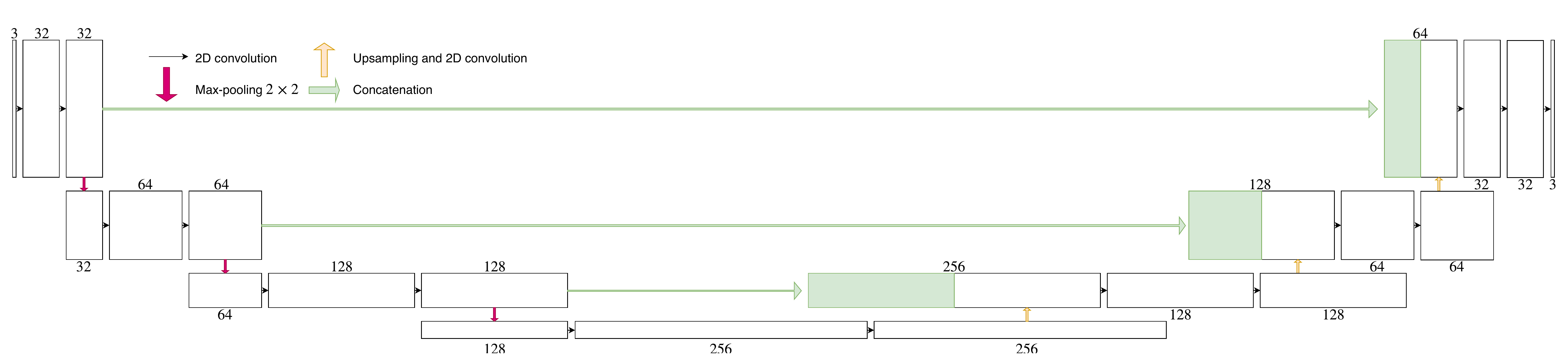}}
	\caption{Architecture of the UDAE proposed model.}
	\label{fig:UDAE}
\end{figure*}
Denoising autoencoder is used for restoring the color of underwater images. We consider the problem of the color restoration as a reconstruction of a corrupted input. Consider that $x$ is the clean image and $\tilde{x}$ is the corrupted version of it by the style transfer $c(\tilde{x}|x)$. Then we would try to reconstruct a repaired input by learning a decoding distribution $p_\theta(x|z)$ from an encoded distribution $q_\phi(z|\tilde{x})$. Denoising autoencoders are expected to capture implicit invariances in the data and extract the key features from the input images~\cite{vincent2008extracting, creswell2018denoising}. U-Net is used as a CNN architecture due to its efficiency in computation and training\footnote{the parameters learn well even with a small dataset} in addition to its ability to propagate context information to higher resolution layers~\cite{ronneberger2015u}.   
 
For a better illustration of the proposed UDAE network, refer to Figure~\ref{fig:UDAE}. Same kernel sizes and layers were used as in UNet~\cite{ronneberger2015u}. First of all, a distorted $RGB$ underwater image is fed into the encoder of the denoising autoencoder. In the encoder part, subsequent convolutions downsample the image gradually to a latent variable. In each downsampling stage, $3 \times 3$ 2-D convolutions are used twice followed by a rectified linear unit (ReLU) and a $2 \times 2$ max-pooling with a stride of $2$. The number of feature maps are doubled in each stage. In the decoder part, upsampling is done from the latent variable back to the original input size sequentially. After each upsampling, the tensor (image) is concatenated with the output of the corresponding symmetric layer in the encoder side and $3$ consecutive convolutions are followed. The feature maps are reduced gradually to $3$ channels. The concatenation of the output of layers combines the contextual information from the downsampling step~\cite{ronneberger2015u}. The reconstructed image should bear resemblance to the clean images, therefore and inspired by the work of Zhao et al.~\cite{zhao2017loss}, Multi-scale Structural SIMilarity (MS-SSIM) index and absolute value ($L1$) loss functions were used. The loss function can be expressed as:
\begin{equation}\label{eq:loss}
\mathcal{L} = \alpha \cdot \mathcal{L}^{MS-SSIM} + (1- \alpha) \cdot \mathcal{L}^{L1},
\end{equation}
where $\mathcal{L}$ represents the loss of the reconstructed image and $\alpha$ is set to $0.80$ after conducting several experiments and observing best reconstruction. The objective of the autoencoder is to minimize the loss function as much as possible. Weight decay is omitted in the proposed network since the presented noise in the input images has a similar regularization effect to weight decay with faster training dynamics~\cite{pretorius2018learning}. Tensorflow framework was used for the training.

\section{Results and Discussion}\label{Results}

\begin{figure*}[htbp]
	\centering
	\subfloat[]{
	\includegraphics[width=80mm,height=25mm]{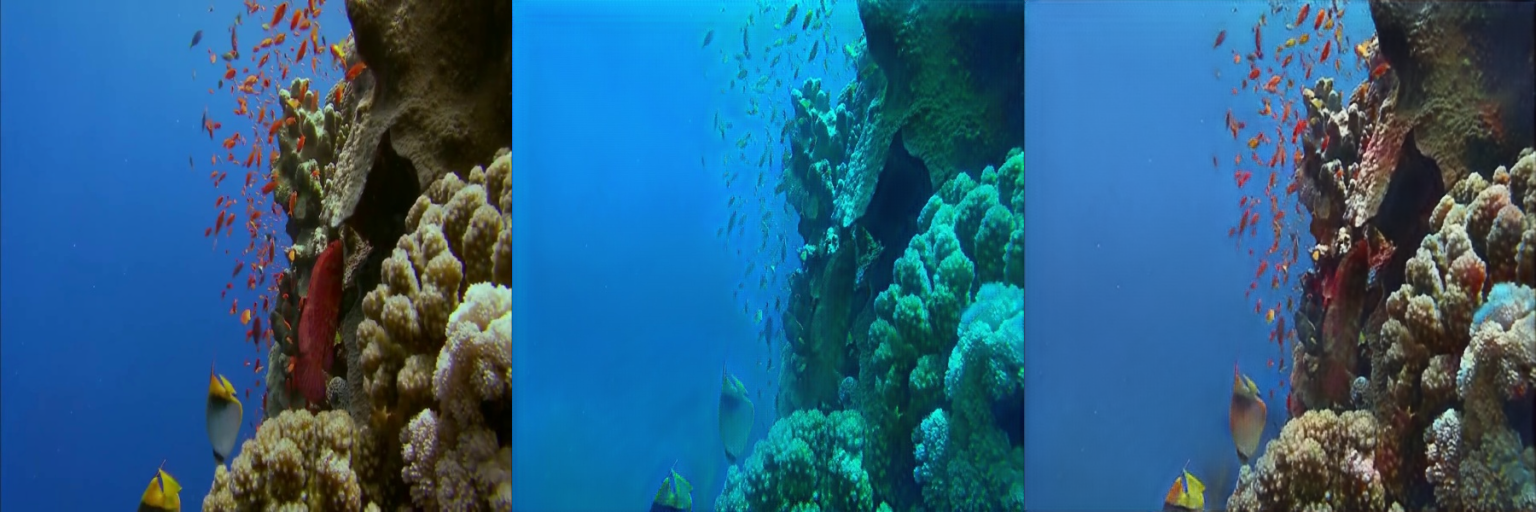}
	}
	\subfloat[]{
	\includegraphics[width=80mm,height=25mm]{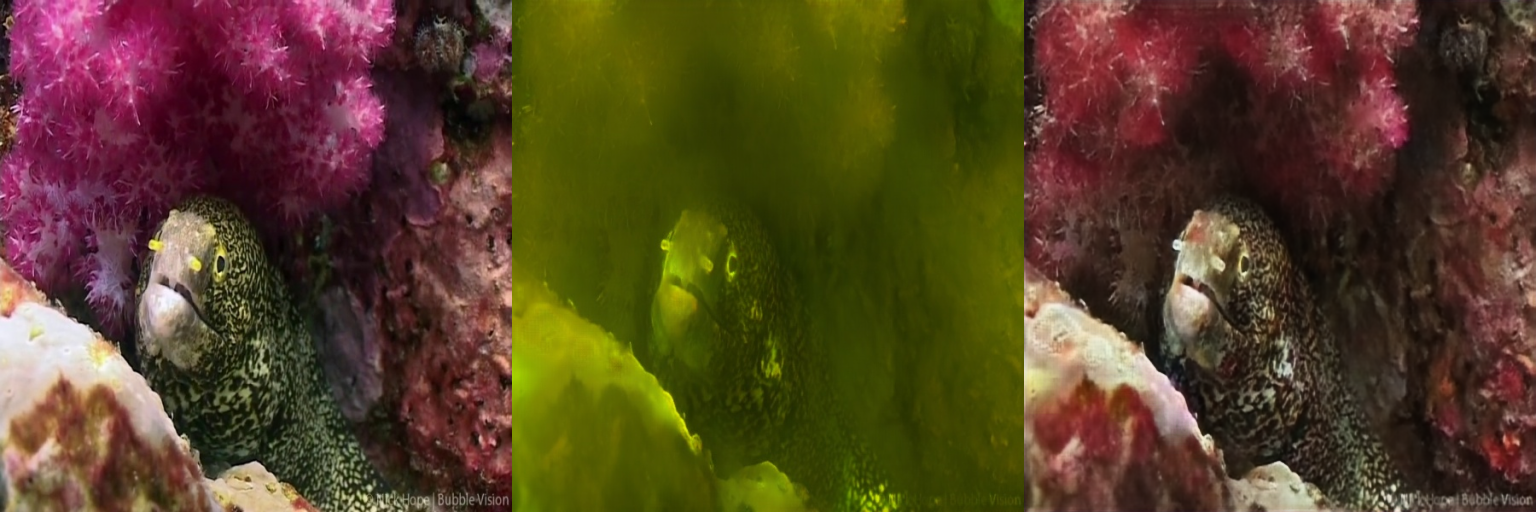}
	}\\
	\subfloat[]{	
	\includegraphics[width=80mm,height=25mm]{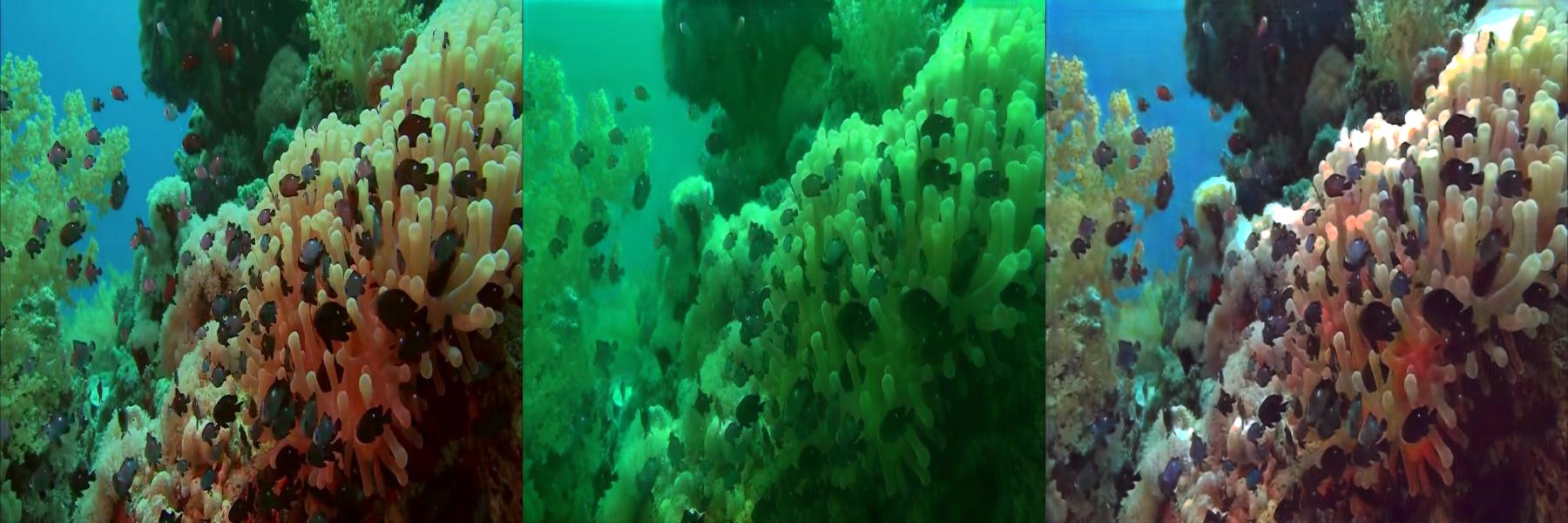}
	\label{subfig:ground-truth-better}	
	}
	\subfloat[]{
	\includegraphics[width=80mm,height=25mm]{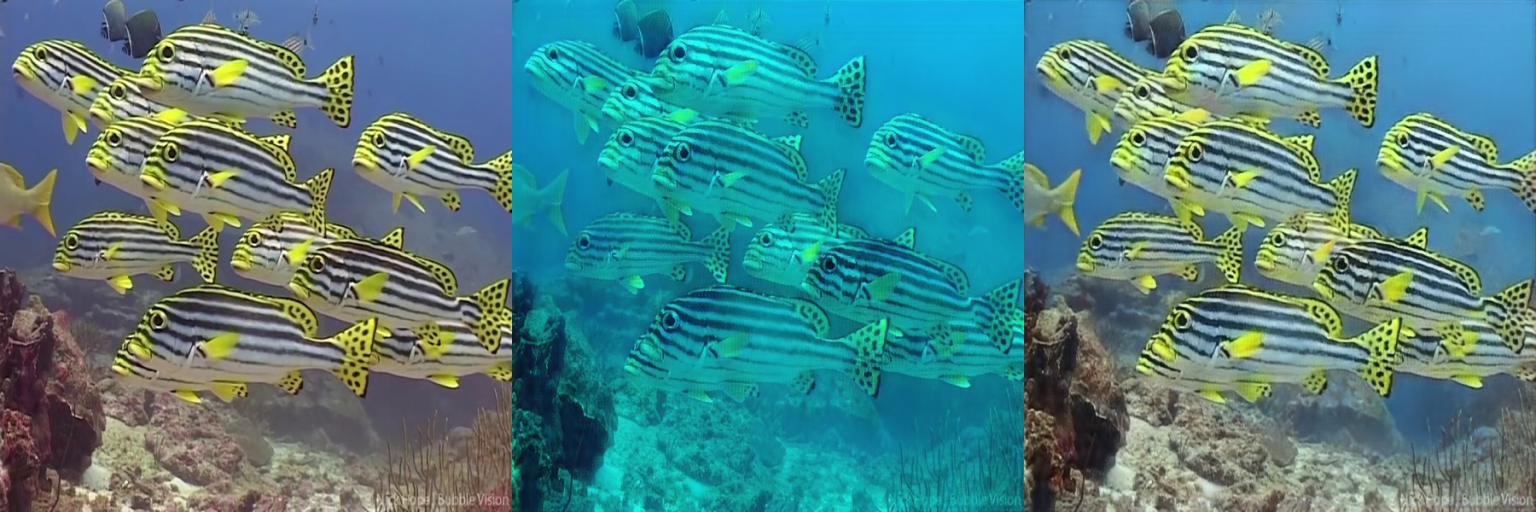}
	}
	\caption{Testing image with its clean and reconstructed version using UDAE. The images from left to right correspond to clean image, input image, and output image respectively.}
	\label{fig:evaluation_yuv}
\end{figure*}

The training of UDAE took around $1$ day on NVIDIA Quadro M5000. It was then tested on $1,040$ images with a resolution of $512 \times 512$. The average time per image in seconds was $0.01601$($62.45$ $fps$) on NVIDIA RTX 2080ti. The selected loss function was capable of preserving details when reconstructing the image. $SSIM$ is sensitive to various types of image degradation~\cite{hore2010image}, whereas $L1$ preserves colors and luminance~\cite{zhao2017loss}. The proposed network produced good results as shown in Figure~\ref{fig:evaluation_yuv} with a suitable speed for real-time implementation.
In certain scenarios where the clean image is only partially clear such as the one in subfigure~\ref{subfig:ground-truth-better}, the reconstructed image showed a better recovery from the distorted color than the clean image itself. The reason is that the network in general learned an encoding and decoding distribution capable of reconstructing color-recovered images.

\begin{figure}[htbp]
	\centering
	\subfloat[Case $1$ - ``Baltic Sea'' Video]{
	\includegraphics[width=40mm,height=20mm]{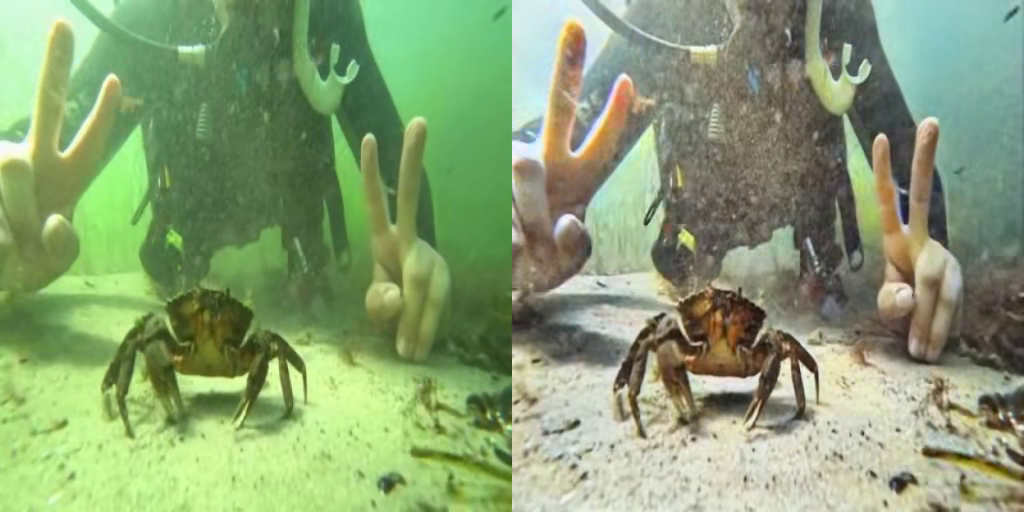}}
	\subfloat[Case $2$ - ``Fish Hunting'' Video]{
	\includegraphics[width=40mm,height=20mm]{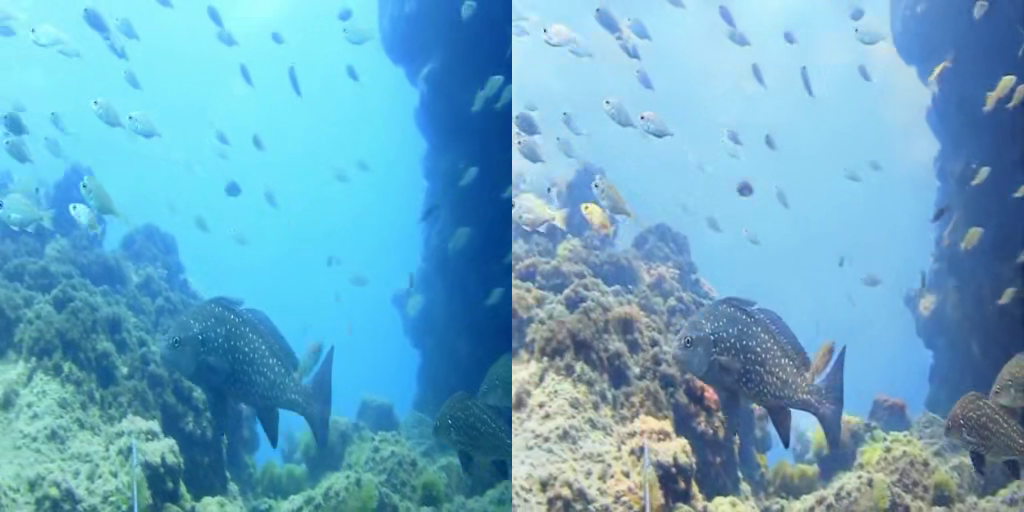}}
    \\
	\subfloat[Case $3$ - ``Baltic Sea'' Video]{
	\includegraphics[width=40mm,height=20mm]{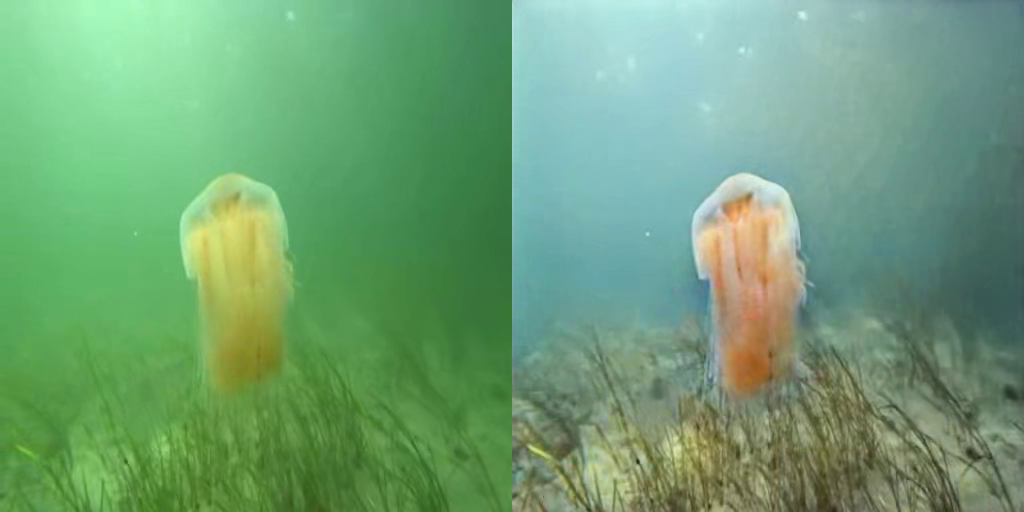}}
	\subfloat[Case $4$ - ``Fish Hunting'' Video]{
    \includegraphics[width=40mm,height=20mm]{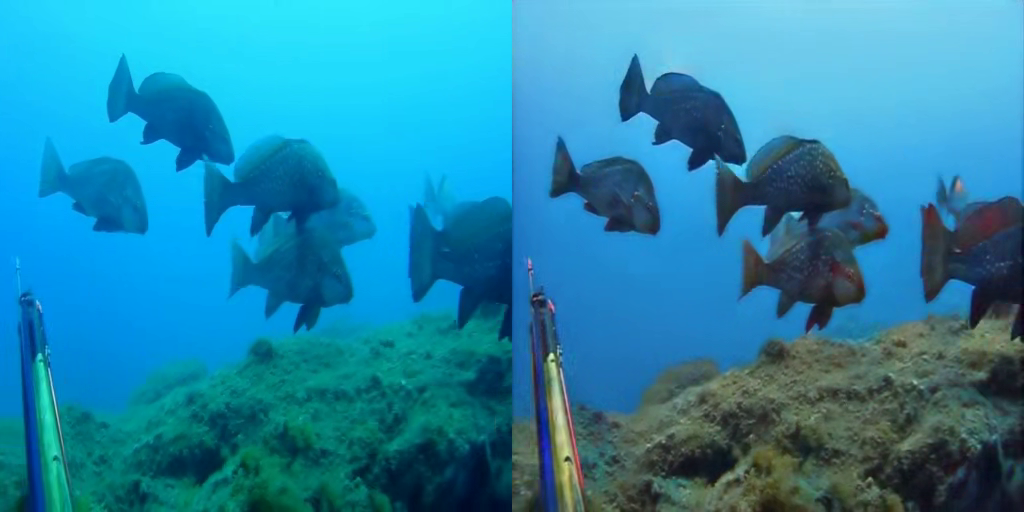}}
    \\
	\subfloat[Case $5$ - ``Scuba Diving'' Video]{
    \includegraphics[width=40mm,height=20mm]{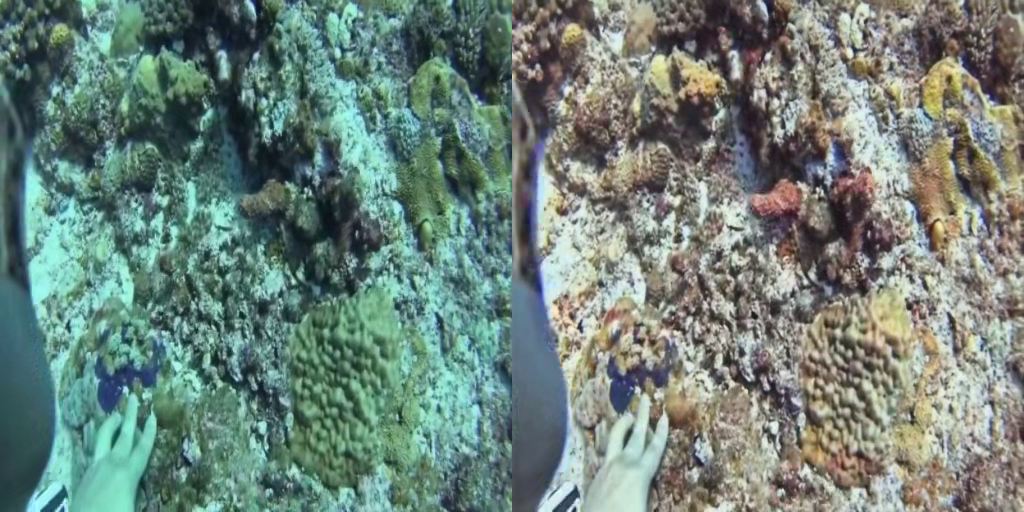}}
	\subfloat[Case $6$ - ``Scuba Diving'' Video]{
    \includegraphics[width=40mm,height=20mm]{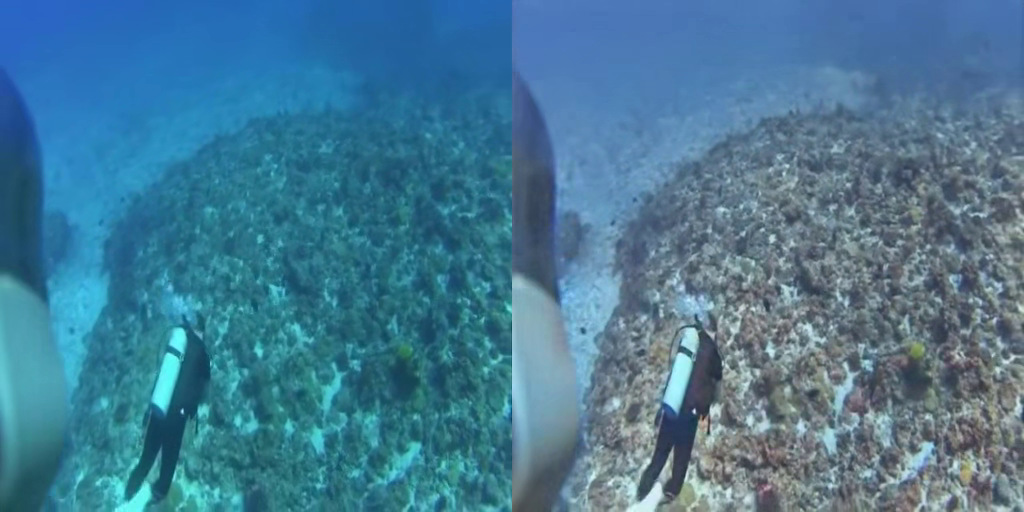}}
	\caption{Samples of the reconstructed frames of the color restoration task extracted from YouTube videos. UDAE performs well on real data even though it was trained on synthetic dataset.}
	\label{fig:color_restoration_videos}
\end{figure}

Additionally, UDAE network was tested on real data such as underwater videos extracted from \href{https://www.youtube.com/}{YouTube} to evaluate its generalization ability. Figure~\ref{fig:color_restoration_videos} shows samples of the reconstructed images on the following videos: \href{https://www.youtube.com/watch?v=Y-SVGO0r6n0}{Baltic Sea}\footnote{https://www.youtube.com/watch?v=Y-SVGO0r6n0}, \href{https://www.youtube.com/watch?v=OSdrb1XNXZI}{Scuba Diving}\footnote{https://www.youtube.com/watch?v=OSdrb1XNXZI}, and \href{https://www.youtube.com/watch?v=aLt7aGFcVkM}{Fish Hunting}\footnote{https://www.youtube.com/watch?v=aLt7aGFcVkM}. The color of the input underwater images with different degradation type was restored and the details were preserved.
\subsection{Comparison with UGAN}
\begin{figure}[htbp]
	\centering
	\subfloat[]{
	\label{goodcolors}
	\includegraphics[width=25mm,height=25mm]{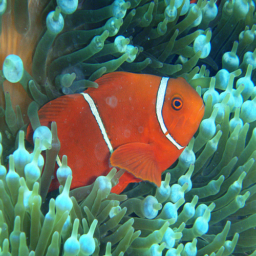}
    \includegraphics[width=25mm,height=25mm]{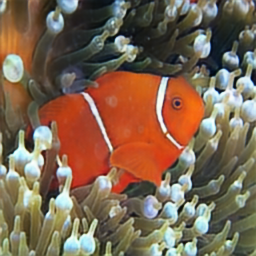}
    \includegraphics[width=25mm,height=25mm]{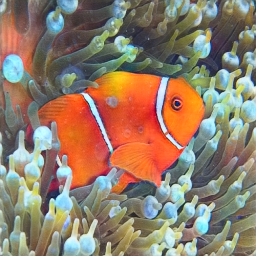}}
    \\
    \subfloat[]{
    \label{goodbackground}
    \includegraphics[width=25mm,height=25mm]{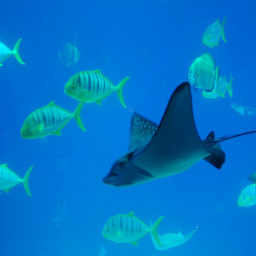}
    \includegraphics[width=25mm,height=25mm]{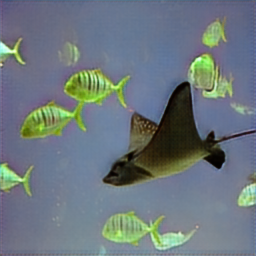}
    \includegraphics[width=25mm,height=25mm]{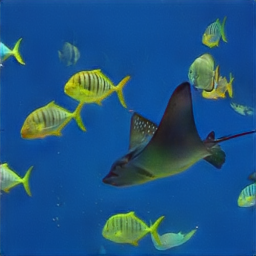}}
    \\
    \subfloat[]{
    \label{gooddetails}
    \includegraphics[width=25mm,height=25mm]{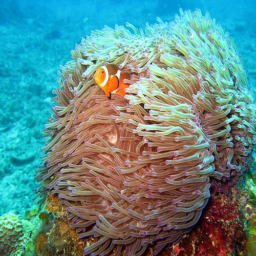}
    \includegraphics[width=25mm,height=25mm]{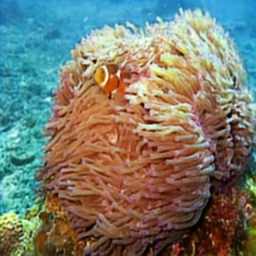}
    \includegraphics[width=25mm,height=25mm]{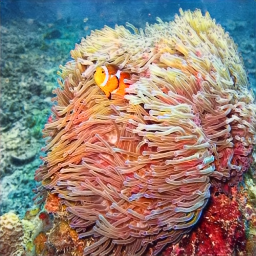}}
    \\
    \subfloat[]{
    \label{badrestoration}
    \includegraphics[width=25mm,height=25mm]{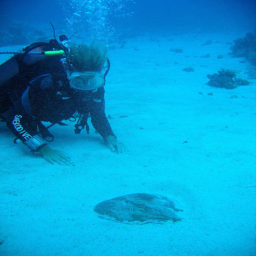}
    \includegraphics[width=25mm,height=25mm]{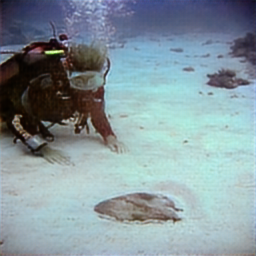}
    \includegraphics[width=25mm,height=25mm]{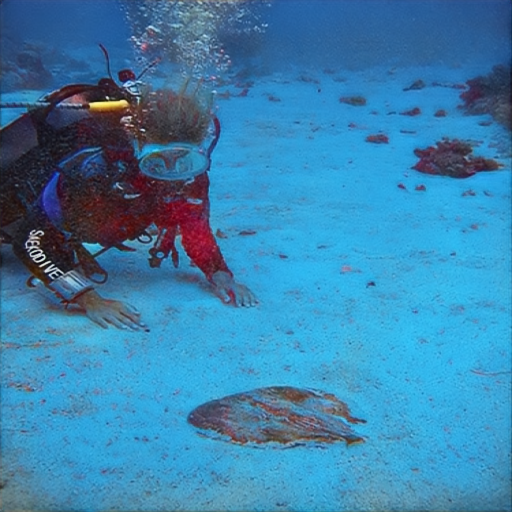}}
	\caption{Comparison to a testing image produced by UGAN. The first column represents the original image, second column represents the reconstructed image using UGAN, and the third column represents the reconstructed image using UDAE.}
	\label{fig:UGAN_UDAE}
\end{figure}

\begin{figure}[htbp]
	\centering
	\subfloat[Input]{
	\includegraphics[width=28mm,height=28mm]{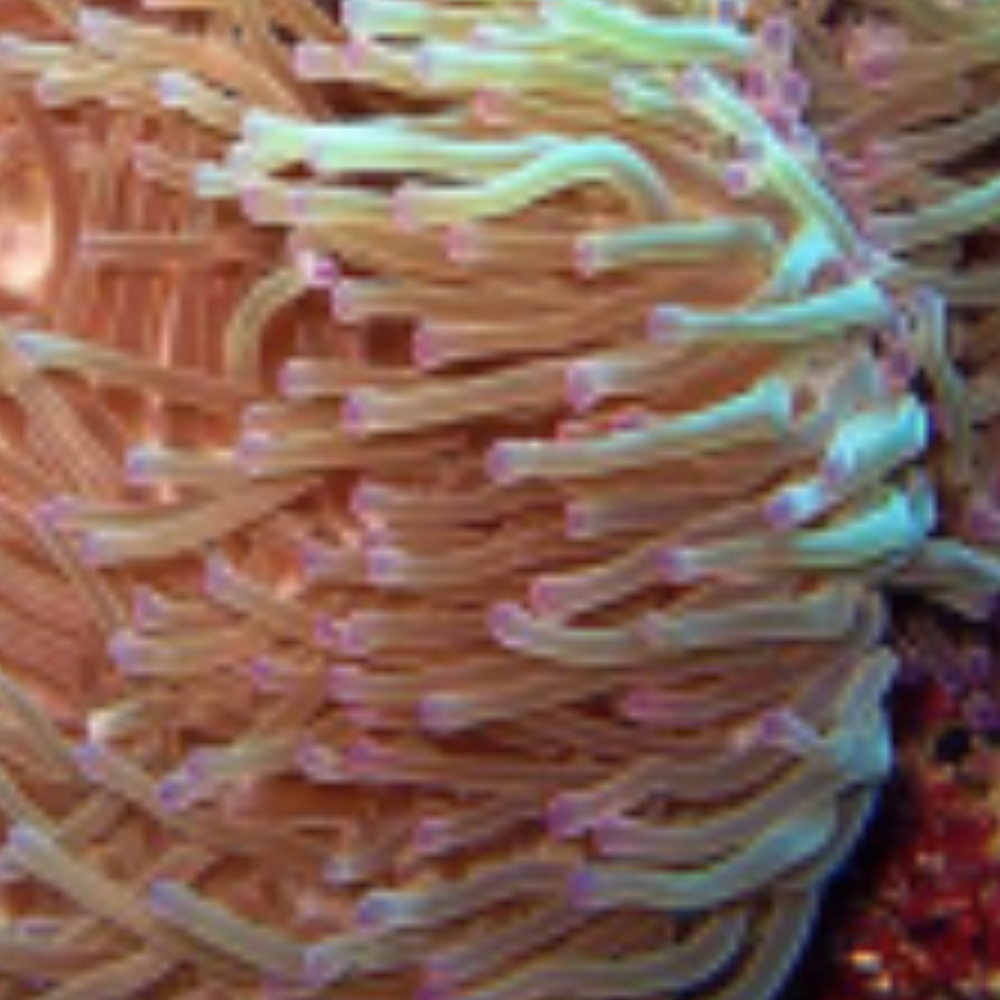}}
	\subfloat[UGAN]{
	\includegraphics[width=28mm,height=28mm]{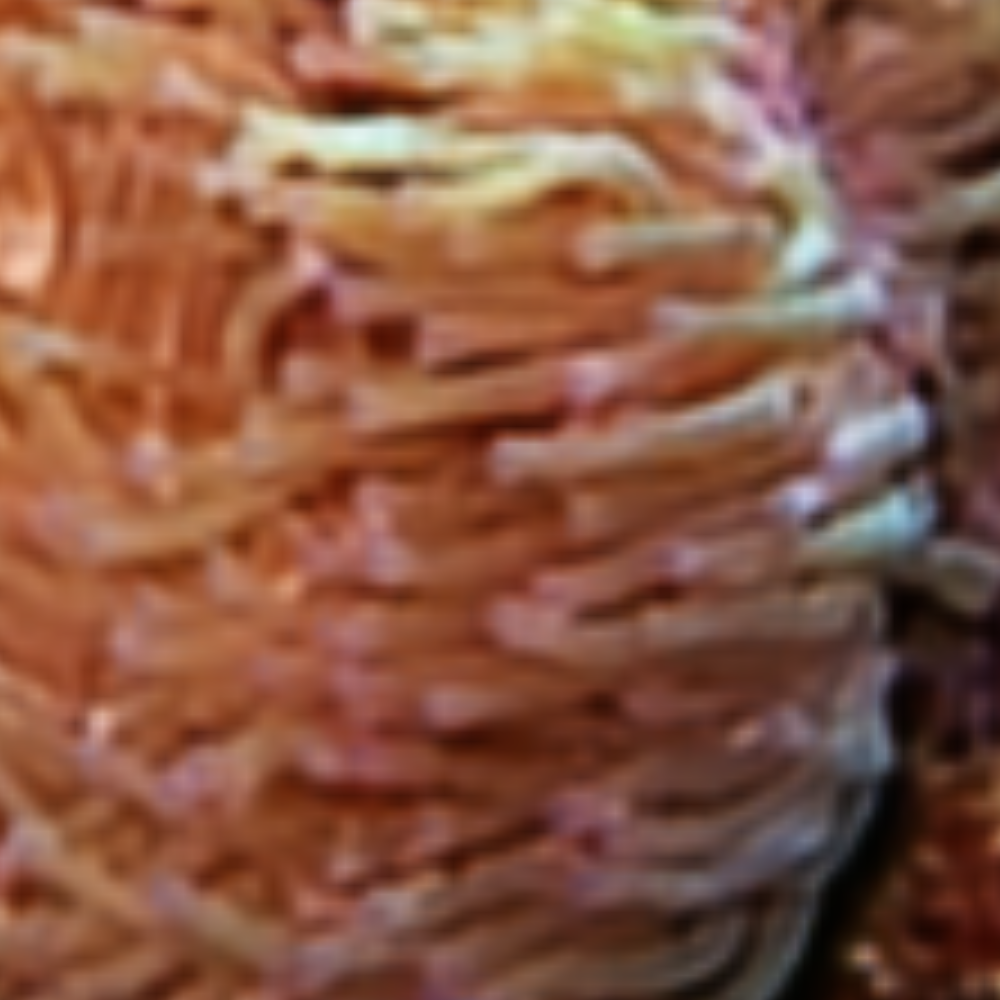}}
	\subfloat[UDAE]{
    \includegraphics[width=28mm,height=28mm]{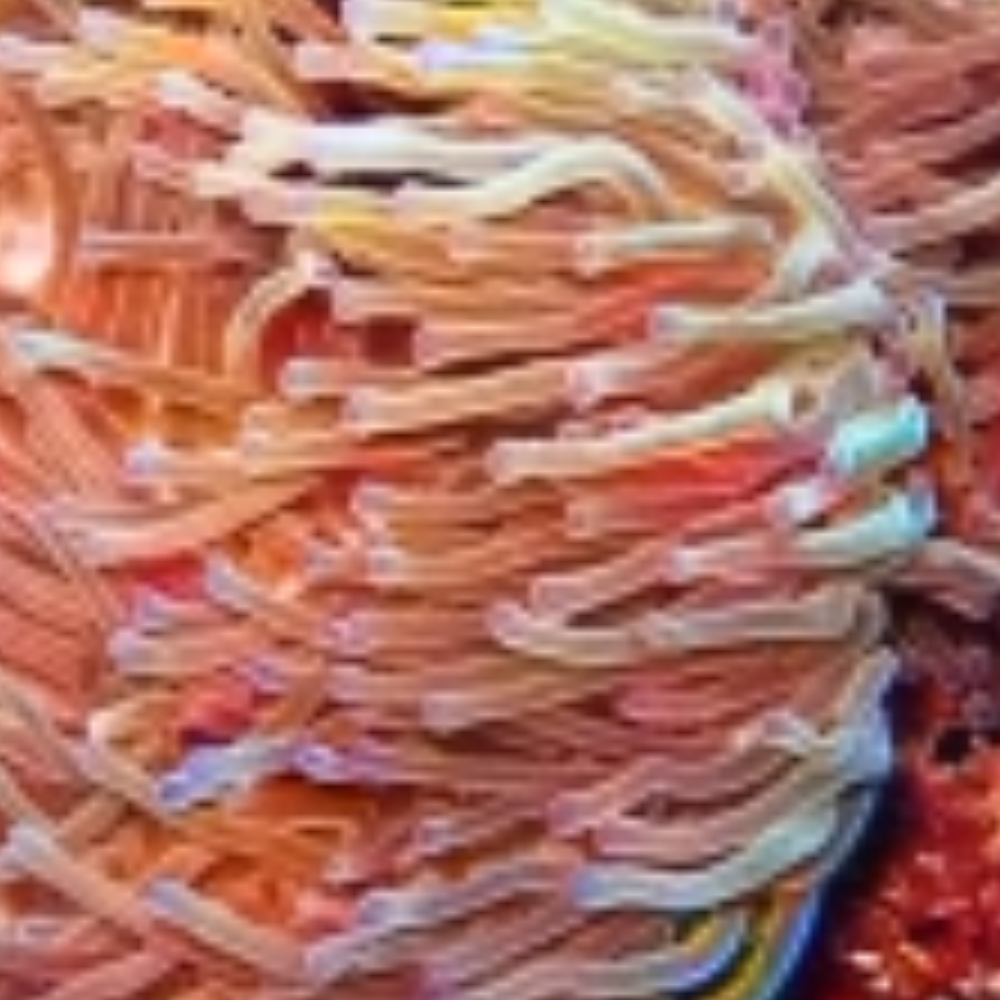}}
	\caption{The input image is zoomed in with a factor of $9$ using bilinear interpolation. The details in the reconstructed image of UDAE are preserved.}
	\label{fig:UGAN_UDAE_zoomed}
\end{figure}

UDAE was compared to Underwater Generative Adversarial Network (UGAN)~\cite{fabbri2018enhancing}. First, both networks were tested on the dataset described in Section \ref{Dataset} ($1,040$ images) due to the availability of the clean image and for an objective evaluation. The testing images are of size $256 \times 256$. Three metrics were used for the evaluation: MSE, SSIM, and MS-SSIM-L1 (eq.~\ref{eq:loss}). In all three metrics, UDAE showed better reconstruction error than that of UGAN, Table~\ref{table:evaluation}\footnote{MSE and MS-SSIM-L1 give a score $0$ for identical images, while SSIM gives a score $1$.}. 
\begin{table}[htbp]
\caption{Evaluation of UGAN and UDAE using three metrics over $1,040$ images with a resolution of $256 \times 256$.}
\begin{center}
\scalebox{1.0}{
\begin{tabular}{|c|c|c|c|}
\hline
\multicolumn{4}{|c|}{\textbf{Objective Evaluation}} \\
\cline{1-4} 
\textbf{Metrics} & \textbf{MSE}& \textbf{SSIM} & \textbf{MS-SSIM-L1}\\
\hline
UDAE& $0.0028$ & $0.9653$ & $0.0753$ \\
\hline
UGAN& $0.0061$ & $0.9186$ & $0.1415$  \\
\hline
\end{tabular}}
\label{table:evaluation}
\end{center}
\end{table}\\
For a fair comparison, both networks were then evaluated on the testing images that the authors of UGAN published in their paper, Figure~\ref{fig:UGAN_UDAE}\footnote{for a better comparison of images, it is better to view them in digital form.}. The average processing time was calculated over $1,813$ testing images resized to $256 \times 256$. The average time per image of UGAN was $0.0099$ seconds ($100.94$ $fps$), whereas that of UDAE was $0.0043$ seconds ($230.67$ $fps$). The processing was conducted on NVIDIA RTX2080ti.Since clean images were not available, the evaluation was only based on the human perception. UDAE showed good generalization where the color was restored and the details were preserved. UGAN achieved good performance in restoring the colors of some images such as subfigure~\ref{goodcolors}, however, UDAE had better color brightness. Another inference drawn from the images is the background reconstruction. In many images, UGAN failed to reconstruct the background properly such as images with plain background as in subfigure~\ref{goodbackground}, whereas our proposed network was capable of restoring the color of both the foreground and the background without any artifacts. An example on the artifacts is the halo effect shown in UGAN reconstructed image. As for the high frequencies, the images of subfigure~\ref{gooddetails} were zoomed in by a factor of $9$ using bilinear interpolation method in Figure~\ref{fig:UGAN_UDAE_zoomed}. 
 UDAE outperforms UGAN network in preserving and reconstructing better details. The coral reefs in the reconstructed image of UGAN were blurry and many details were lost. The details are important for object detection and tracking by underwater vehicles. Some failure cases were noticed by our proposed network such as subfigure~\ref{badrestoration}. This will be kept for future work where a better dataset with more degradation types would be established for a better generalization. 

\section{Conclusion}\label{Conclusion}

This paper proposed Underwater Denoising Autoencoder (UDAE); a new way for restoring the color of underwater images using a single denoising autoencoder with real-time capability. We showed that it is possible to reconstruct underwater images using a network based on a single denoising autoencoder, where it gave same or better results than a network based on a GAN. However, using a single autoencoder is better suited for real-time implementation. Additionally, as an improvement to previous networks, UDAE is capable of restoring better color in images and preserving the details.

We believe that there is a space for improving the network, where better generalization ability should be achieved. The network was trained on a relatively small dataset, however, obtaining a larger one with various color distortion would lead to great improvement. The processing speed could be also improved by trying different CNN-baseline or latent space size. 

\bibliographystyle{./IEEEtran}
\bibliography{./Bibliography}
\end{document}